\documentclass[letterpaper]{article} 
\usepackage{amsmath}
\usepackage{amsfonts}
\usepackage{subcaption}
\usepackage{aaai25}  
\usepackage{times}  
\usepackage{helvet}  
\usepackage{courier}  
\usepackage[hyphens]{url}  
\usepackage{graphicx} 
\usepackage{natbib}  
\usepackage{caption} 
\usepackage{pifont}
\newcommand{\xmark}{\ding{55}}%
\usepackage{booktabs}    
\usepackage{caption}     

\usepackage{makecell}
\usepackage{colortbl}
\usepackage{algorithm}
\usepackage{algorithmic}
\usepackage{newfloat}
\usepackage{listings}
\DeclareCaptionStyle{ruled}{labelfont=normalfont,labelsep=colon,strut=off} 
\floatstyle{ruled}
\newfloat{listing}{tb}{lst}{}
\floatname{listing}{Listing}
\title{MCAT: Visual Query-Based Localization of Standard Anatomical Clips in Fetal Ultrasound Videos Using Multi-Tier Class-Aware Token Transformer}

\author{
    Divyanshu Mishra\textsuperscript{\rm1}, Pramit Saha\textsuperscript{\rm1}, He Zhao\textsuperscript{\rm3}, Netzahualcoyotl Hernandez-Cruz\textsuperscript{\rm1}, Olga Patey\textsuperscript{\rm2}, Aris Papageorghiou\textsuperscript{\rm2}, J. Alison Noble\textsuperscript{\rm1}}

\affiliations{
\textsuperscript{\rm1}Department of Engineering Science, University of Oxford\\
    \textsuperscript{\rm2}Nuffield Department of Women's and Reproductive Health, University of Oxford\\
    \textsuperscript{\rm3}Institute of Life Course and Medical Sciences, University of Liverpool\\
    
    \{divyanshu.mishra, pramit.saha, netzahualcoyotl.hernandez-cruz, alison.noble\}@eng.ox.ac.uk,\\
    \{olga.patey, aris.papageorghiou\}@wrh.ox.ac.uk, he.zhao@liverpool.ac.uk
}

\usepackage{bibentry}

\begin{document}
\maketitle
\begin{abstract}
Accurate standard plane acquisition in fetal ultrasound (US) videos is crucial for fetal growth assessment, anomaly detection, and adherence to clinical guidelines. However, manually selecting standard frames is time-consuming and prone to intra- and inter-sonographer variability. Existing methods primarily rely on image-based approaches that capture standard frames and then classify the input frames across different anatomies. This ignores the dynamic nature of video acquisition and its interpretation.
To address these challenges, we introduce Multi-Tier Class-Aware Token Transformer (MCAT); a visual query-based video clip localization (VQ-VCL) method to assist sonographers by enabling them to capture a quick US sweep. By then providing a visual query of the anatomy they wish to analyze, MCAT returns the video clip containing the standard frames for that anatomy, facilitating thorough screening for potential anomalies.
We evaluate MCAT on two ultrasound video datasets and a natural image VQ-VCL dataset based on Ego4D. Our model outperforms state-of-the-art methods by 10\% and 13\% mtIoU on the ultrasound datasets and by 5.35\% mtIoU on the Ego4D dataset, using 96\% fewer tokens.
MCAT’s efficiency and accuracy have significant potential implications for public health, especially in low- and middle-income countries (LMICs), where it may enhance prenatal care by streamlining standard plane acquisition, simplifying US based screening, diagnosis and allowing sonographers to examine more patients. 

\end{abstract}

\section{Introduction}
\label{sec:introduction}

\begin{figure}
    \centering
\includegraphics[width=1\columnwidth]{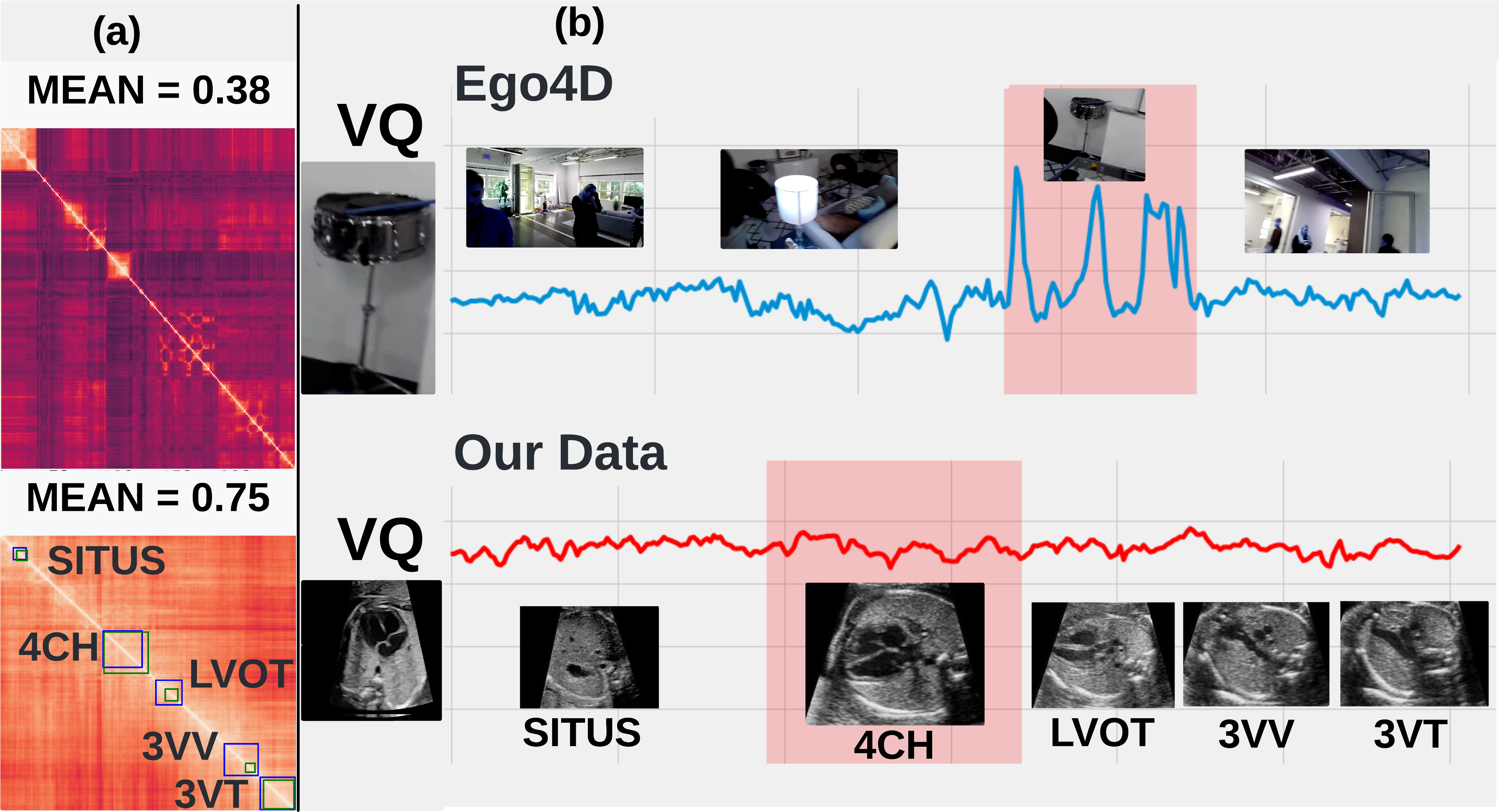}
    \caption{(a) Self-similarity matrix for a randomly chosen video from Ego4D (\textbf{top}, mean=0.3) \cite{grauman2022ego4d} and our clinical video dataset (\textbf{bottom}, mean=0.75), which reveals higher task difficulty for our video clip localization task. 
    The uncertainty in the annotations of two expert cardiologists is shown in green and blue boxes.
    (b) Cosine similarity of the visual query with the video for both Ego4D (top) and our data (bottom). 
    Our clinical data obtains similar scores along the video 
    emphasizing the challenge, whereas Ego4D exhibits high scores only within region of interest.}
\label{fig1}
\end{figure}

Fetal ultrasound is essential for monitoring prenatal development, detecting potential abnormalities, and ensuring the health of both the fetus and the expectant mother. In routine pregnancy assessments, a sonographer scans different fetal anatomies to assess fetal development and identify anomalies. Selecting standard frames \cite{salomon2022isuog,hernandez2025comprehensive,mishra2023dual} that meet clinical guidelines (\textit{e.g.}, ISUOG) is a time-consuming process, and a typical fetal ultrasound scan can take up to an hour. 
Multiple studies have attempted to streamline this process by automatically identifying standard planes in 2D fetal ultrasound using deep learning-based classification models \cite{rahmatullah2011automated, cai2018multi, lee2021principled, baumgartner2016real, schlemper2018attention}. Other recent works have looked into leveraging temporal information for more complex tasks, such as anomaly detection in ultrasound videos \cite{zhao2022towards,zhao2023memory} and generative modeling of standard planes \cite{men2023towards}, although these approaches do not explicitly localize standard frames. 
Integrating a video-clip localization model could enhance sonographer workflow by allowing them to focus on detailed video reviews and anomaly detection. However, automatically detecting standard frames in video is challenging due to the high similarity of frames before and after the standard ones, making it difficult to determine temporal anatomical boundaries, as shown in Fig. \ref{fig1}. Additionally, even human experts may find it hard to agree on standard frame selection, as evidenced by our study showing a kappa score of only 66\% between two fetal cardiologists annotating the same fetal heart videos (see Supp. Fig. 1), highlighting the complexity and inherent noise in annotations.
Text query-based localization tasks, such as video-temporal grounding \cite{yang2022tubedetr,zeng2020dense,lin2023univtg}, video moment retrieval \cite{liu2022umt,moon2023query}, and highlight detection, have shown promising performance in natural video understanding. However, textual data often falls short of providing the dense video understanding required for some applications. In the medical domain, reports traditionally rely on static images and text to convey diagnostic information \cite{moon2022multi,saha2024examining,saha2024fedpia,saha2024f}. While image-based methods are informative, video-based analysis can offer a significant advancement in diagnostic capabilities. For instance, a dynamic ultrasound video of a beating heart provides a more detailed and holistic assessment of cardiac function compared to a single static frame \cite{scott2013increasing}. Similarly, in fetal anatomy examinations, video clips allow practitioners to measure biometry more accurately and review the entire sequence for optimal plane selection, thereby enhancing diagnostic precision.
Despite the advantages, paired video-textual data is typically scarce in the medical field. When available, it usually includes sparse class labels or radiology reports providing a diagnosis for the entire video rather than detailed clip-level information. This is where image-based queries, or visual queries (VQs), are potentially valuable. VQs allow for intuitive and direct identification of objects or similar images, reducing language barriers and effectively expressing complex concepts that might be difficult to articulate through text. For example, describing a medical anomaly can be challenging with a text query, whereas an example frame containing the anomaly can provide a more effective query for model training. In the context of ultrasound videos, retrieving a video clip rather than a single frame is more challenging due to the motion of the ultrasound probe and the scanned object, leading to various deformations, occlusions, and motion blur, making it harder to localize all instances of the object as shown in Fig. \ref{fig1}.
\noindent To reduce the time to conduct a full scan assessment, we introduce the Visual-Query-based Video Clip Localization (VQ-VCL) task. In this approach, a sonographer performs a quick sweep to capture all relevant anatomies. With a visual query depicting the required anatomy, our method can automatically select the relevant standard-frame clips from this video sweep. This significantly reduces manual effort, enhances efficiency, and allows sonographers to scan more patients while focusing more on analyzing the standard video clips. 

To tackle the challenges of the VQ-VCL task, we introduce MCAT, a Multi-Tier transformer-based model with class-specific tokens. It consists of three primary components: a Multi-Tier Class-Aware Spatio-Temporal Transformer for modeling spatial and temporal interactions and learning class-specific features through class-specific tokens, a Temporal Uncertainty Localization Loss to mitigate label noise, and a Multi-Tier, Dual Anchor Contrastive Loss for addressing complex event boundaries.

Our contributions are as follows:
\begin{enumerate}
    \item We introduce the VQ-VCL task and propose MCAT, a spatio-temporal video Transformer model for automatic standard-plane video clip retrieval.
    
    \item We propose a multi-tier feature extraction module to learn spatio-temporal features in a coarse-to-fine manner. A query-aware Transformer captures spatial information, while temporal information is condensed into class-specific learnable tokens. These tokens disentangle class-specific features into distinct tokens, improving video clip localization and significantly boosting model efficiency by reducing the number of tokens by 96\%. This makes the approach potentially suitable for applications in resource-constrained public health including low- and middle-income country (LMIC) settings.

    \item We propose a hybrid loss function comprising Multi-Tier, Dual Anchor Contrastive Loss, and Temporal Uncertainty-Aware Localization Loss to handle complex event boundaries and noisy labels.
    
   \item We assess model performance on two real-world clinical datasets for standard-plane detection with limited data and annotations which naturally contain a high degree of noise. Additionally, we create and evaluate our model on an open-source VQ-VCL natural videos dataset based on Ego4D \cite{grauman2022ego4d}.
\end{enumerate}

\begin{figure*}[t]
    \centering
    \includegraphics[width=0.9\textwidth]{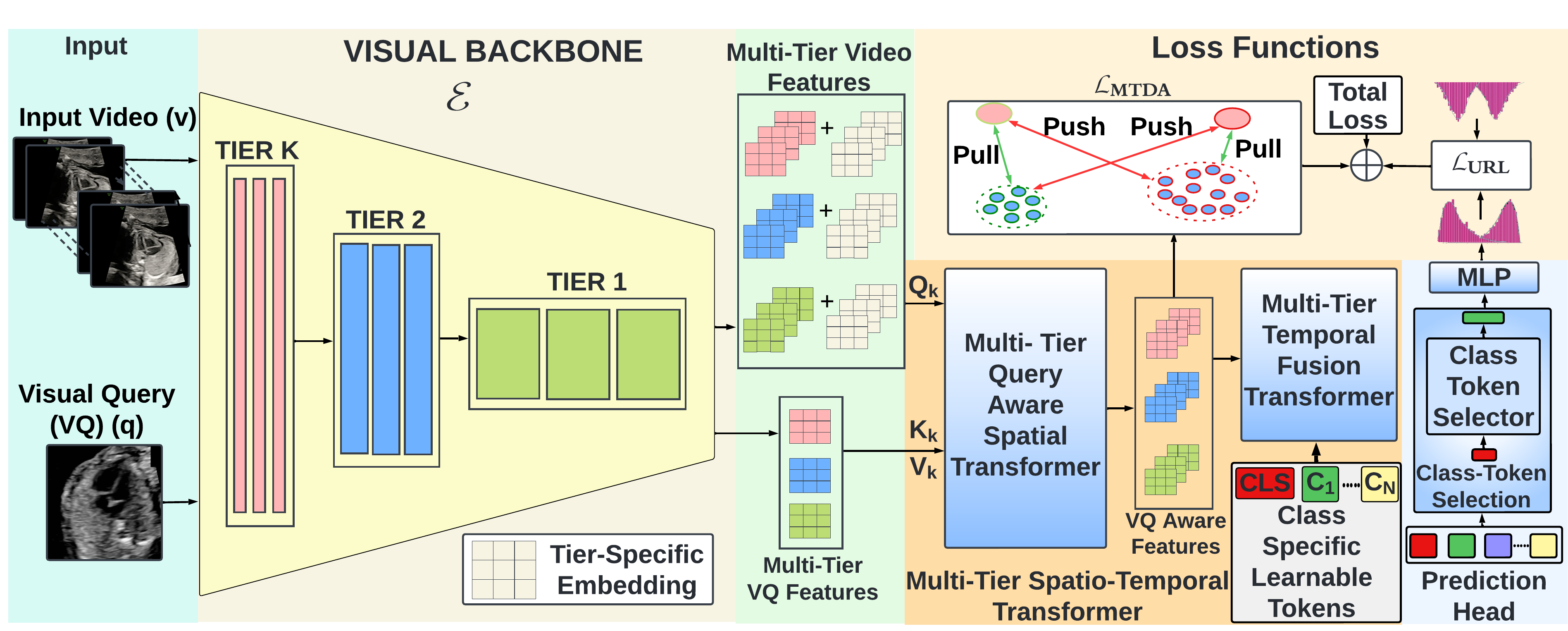}
    \caption{Main architecture of MCAT. The input video $v$ and visual query $q$ are passed to the visual backbone to give multi-Tier features. These features are fused spatially using the Multi-Tier Query Aware Spatial Transformer. The Tier-specific features are passed to a) ${\mathcal{L}_{MTDA}}$ to learn the separation between classes, b) the Multi-Tier Temporal Fusion transformer to learn Tier-Aware Spatio-Temporal Embedding, which is further passed to an MLP to make final prediction and calculate $\mathcal{L}_{URL}$ loss.}
    \label{fig2}
\end{figure*}
\section{Methods}

\begin{figure}[t]
    \centering
    \includegraphics[width=1\columnwidth]{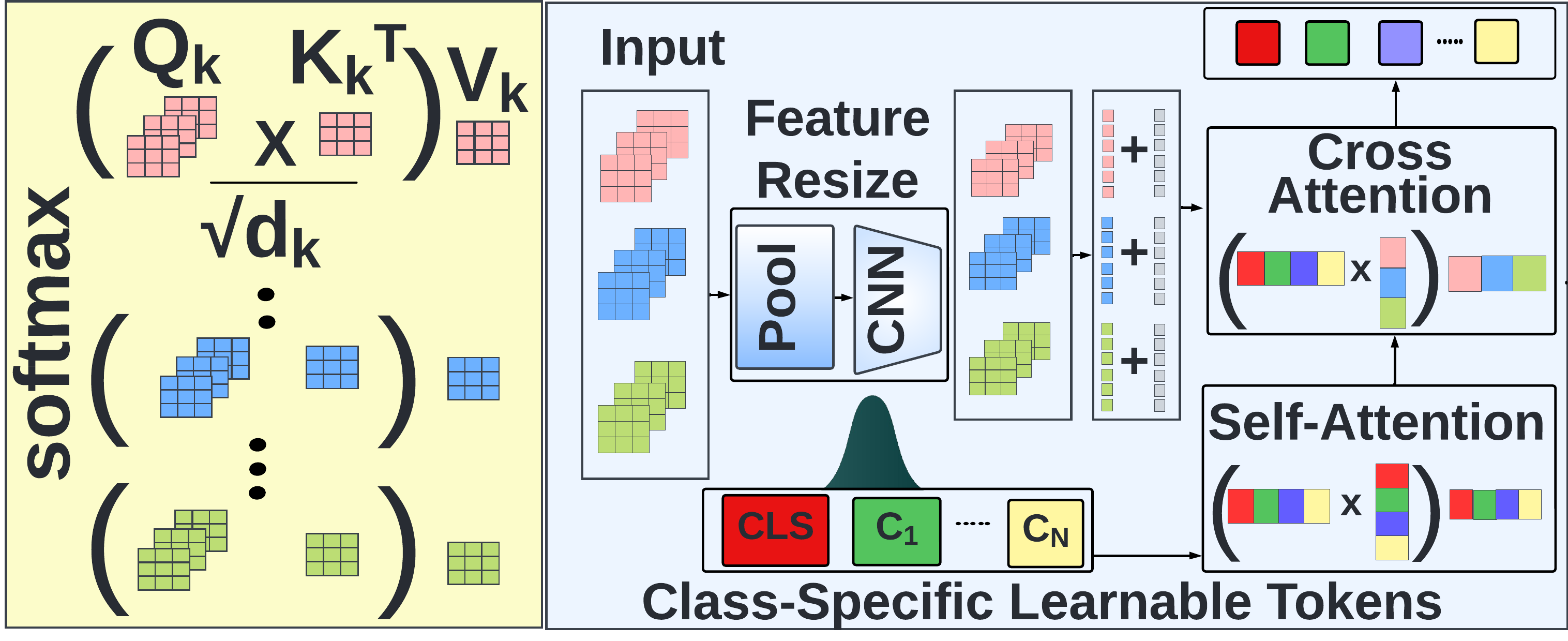}
    \caption{Fig (left) shows the spatial feature fusion mechanism where Tier-specific video and VQ features are spatially fused to give Tier-specific query-aware features. Figure 3 (right) shows how the Tier-specific query-aware features are first resized, flattened and enriched with positional information. The resulting features are concatenated and fused to learn the Tier-Aware Spatio-Temporal Embedding.}
    \label{fig3}
\end{figure}
\subsection{Video Clip Localization Task Formulation}
\noindent The visual query-based video clip localization (VQ-VCL) task is formulated as a temporal localization task. Formally, given a video ${v}$ and an exemplar frame ${q}$ from a separate exemplar database $\mathcal{Q}$, the model is trained to predict the start (${t_s}$) and end (${t_e}$) frame number of a  clip ${{v_q}}$  where ${{v_q}} \subset {v}$ and contains frames semantically similar to $q$.
\subsection{MCAT Overall Architecture}
\noindent Our proposed model, illustrated in Fig. \ref{fig2}, processes a video \(v\) and a visual query \(q\) as inputs. These inputs are fed through a shared encoder \(\mathcal{E}\), generating \(K\) tier video features \(f_{{v}_K} \in \mathbb{R}^{T \times H_k \times W_k \times C_k}\) and visual query features \(f_{{q}_k} \in \mathbb{R}^{H_k \times W_k \times C_k}\), where \(k\) iterates over the \(K\) tiers and \(T\), \(H_k\), \(W_k\), and \(C_k\) represent the number of frames, height, width, and channel dimensions at tier \(k\). 
The features extracted at each tier from a visual backbone \(\mathcal{E}\) are enriched with scale-aware learnable embeddings and spatially fused using a Multi-Tier Query-Guided Spatial Transformer, resulting in multi-tier VQ aware features as shown in Fig. \ref{fig2}. 
A multi-tier temporal fusion transformer is proposed to learn a series of tokens (CLS, $\{ 
C_i | i=1,\cdots, N\}$) from the VQ aware features where N is the number of classes. 
The tokens are further utilized to predict the start and end frames.


\subsection{Multi-Tier Spatio-Temporal Transformer}
\subsubsection{Multi-Tier Query Guided Spatial Transformer }
The design of the encoder to fuse the video and visual query features is crucial, especially in fine-grained video localization settings where the classes are highly similar. Previous work on visual grounding \cite{yang2022tubedetr} and moment retrieval \cite{lei2021detecting} naively concatenates the features from the video and query together. This approach can diminish the relevance of visual queries and result in features with low information about the visual query \cite{moon2023query}. Moreover, these works are designed for text query-based video retrieval where modality features are only extracted in a single hierarchy. In contrast, features from videos and images can be extracted at multiple tiers, each tier containing coarse to fine-grained information. This variability in information can be beneficial for retrieval, especially in scenarios where the classes are highly similar with some local variations.
To ensure video features at Tier \(k\) (\(f_{{v}_k}\)) are contextualized by visual query features (\(f_{{q}_k}\)) from the respective tier, we designed a Multi-Tier Query Guided Spatial Transformer where \(k = 1, 2, 3, \ldots, K\). We achieve this by extracting features from \(K\) tiers of the shared visual backbone for the video and the visual query. Tier-specific learnable embeddings (\(emb_K\)) are added to each tier video feature to ensure optimal learning and fusion of tier-specific information from the video and visual query. 
The resulting video features and visual query features for each tier are then fused using cross-attention \cite{vaswani2017attention} to learn these tier-specific embeddings (\(emb_K\)). Formally, given the video feature \(f_{{v}_k}\) and visual query feature \(f_{{q}_k}\) at tier \(k\) where \(k = 1, 2, 3, \ldots, K\) and \(k = 1\) means the features from the last layer of the visual backbone. We first add to each tier video feature the tier-specific learnable embedding (\(emb_K\)) such that \(f_{{v}_k} = f_{{v}_k} + emb_k\). We project the video feature to get query (\(Q_{v_k}\)), whereas key (\(K_{q_k}\)) and value (\(V_{q_k}\)) are obtained from the visual query feature. The attention mechanism \cite{vaswani2017attention} is applied to \(Q_{v_k}\), \(K_{q_k}\), and \(V_{q_k}\), and the output is passed to a feed-forward network as shown in Eq. \ref{eq1} to produce the tier-specific query-aware video features (\(QV_{f_k}\)) for tier \(k\). This process is performed in parallel for all \(K\) tiers to obtain tier-specific query-aware features $QV_{f_K}$ for each tier.
\begin{small}

\begin{align}
\label{eq1}
    QV_{f_K} &= FFN\left(\text{softmax}\left(\frac{Q_{v_k}{{K}_{q_k}^T}}{\sqrt{d_k}}\right)V_{q_k}\right)
\end{align}
\end{small}
\subsubsection{Multi-Tier Temporal Fusion Transformer }
To incorporate temporal information into the Tier-specific query-aware video features ${QV_{f_{_k}}}$ and to fuse these spatio-temporal features across Tiers for learning the class-specific Tier-aware spatio-temporal tokens (CLS, $\{ 
C_i | i=1,\cdots, N\}$) where N is number of classes,
we designed a multi-Tier temporal fusion transformer.
Formally, given the Tier-specific query-aware features for each Tier ${QV_{f_{_k}}}$, and randomly initialized class-selection token $CLS$, Class-Specific Tier-Aware Spatio-Temporal learnable tokens ${C}_{N} \in $ $\mathbf{R^{(N) \times {H_M} \times {W_M} \times {C_M}}}$  where $k$= $1,2,3 \ldots K.$ 
We first perform self-attention between the ${E_T} = CLS + {C}_{N}$ tokens as in Eq. \ref{eq2}.
\begin{equation}
\label{eq2}
    {E}_{T} = FFN\left(\text{softmax}\left(\frac{Q_v\left({K_{v}^T}\right)}{\sqrt{d_k}}\right)V_v\right)
\end{equation}
Cross-attention is performed between the resulting vector and the Tier-specific query-aware video features ${QV_{f_{_k}}}$ as formulated in Eq.\ref{eq2.1} and shown in Supp. Fig 3.
\begin{equation}
\small
\label{eq2.1}
    {E}_{T} = FFN\left(\text{softmax}\left(\frac{Q_{E_T}\left({K_{QV_{f}}^T}\right)}{\sqrt{d_k}}\right)V_{{QV}_{f}}^T\right)
\end{equation}

This helps fuse the spatial and temporal information available across the Tiers into the class-specific Tier-Aware Spatio-Temporal tokens. The spatio-temporal information-rich $E_T$ tokens are fed to the token selection block that helps select the token corresponding to the VQ and updates only the selected token with the current spatio-temporal class-specific information.
\subsubsection{Class-Specific Token Selection and Learning}
The block is designed to select the class-specific token (\(C_S\)) corresponding to the visual query and to enable class-specific token learning. During training, the class-selection token (CLS) obtained after spatio-temporal fusion is passed through a multi-layer perceptron (MLP) to predict the class to which the visual query (VQ) belongs. This is formulated as an \(N\)-class classification problem, and cross-entropy loss is used to train the MLP. Since the class of the VQ is known during training, we use this information to select the class-specific token (\(C_S\)) and only update the token for the specific VQ class. During inference, the prediction from the trained MLP is used to select \(C_S\) and predict the start and end frames of the ground-truth video clip.

\subsection{Loss Functions}
\label{sec3.4}
\subsubsection{Multi-Tier, Dual Anchor Contrastive Loss}
In settings with high spatial similarity between the video frames, as seen in Fig. \ref{fig1}, estimating the correct event boundary is challenging. Moreover, in such a case, object appearance can significantly vary from that of the visual query as the objects of interest and the data acquisition device are both in motion.
To mitigate the above challenges and learn subtle differences between the classes, we propose a Multi-Tier, Dual Anchor Contrastive Loss, where the anchors and samples are selected from different tiers. The loss function has two main components: 
1. \textbf{Multi-Tier Positive Anchor Contrastive Loss ($L_{PAC}$)}, which aims to bring the tier-specific visual query-aware features in the ground-truth clip together while pushing away features belonging to other classes.
2. \textbf{Multi-Tier Negative Anchor Contrastive Loss ($L_{NAC}$)}, which utilizes a negative anchor to further push the positive tier-specific query-aware features away from the negative ones.
Formally, given Tier-Specific Query Aware features \( {f_{{vq}_k}} \) for each tier, we project the features to a shared feature space to ensure that only rich-semantic features from each tier are captured. This is achieved through a CNN projection layer \( {P_\theta}_{k} \), resulting in projected features \( {f_{{vq}_k}'} \). Subsequently, we extract the video features belonging to the ground-truth clip and define them as positive features (\( {f'}_{vq_k}^+ \)) for each tier. The video features of the frames lying outside the ground-truth clip are defined as negative features (\( {f'}_{vq_k}^- \)).
We randomly sample a tier and utilize its features as anchors. A tier's positive features serve as the positive  (\( {f}_{vq_a}^+ \)), while the negative features serve as the negative anchor (\( {f'}_{vq_a}^- \)) for the remaining tiers. For the Positive Anchor Contrastive Loss, we calculate the cosine similarity between (\( {f'}_{vq_a}^+ \), \( {f'}_{vq_k,_i}^+ \)) and (\( {f'}_{vq_a}^+ \), \( {f'}_{vq_k,_j}^- \)) as stated in Eq. \ref{eq5}, where \(\text{sim}(.)\) denotes the cosine similarity function and \(i, j\) iterate over \(M_1\) positive and \(M_2\) negative samples, while \(k\) iterates over \(K-1\) tiers.

\begin{equation}
\small
\label{eq5}
\mathcal{{L}}_{PAC}=-\log \frac{\sum_{k=1}^{K-1} \sum_{i=1}^{M_1} \exp \left(sim({f'}_{vq_a}^+,{f'}_{vq_k,_i}^+)/ \tau^+\right)}{\sum_{k=1}^{K-1} \sum_{j= 1}^{M_2} \exp \left(sim({f'}_{vq_a}^+,{f'}_{vq_k,_j}^-)/ \tau^+\right)}
\end{equation}

\
Finally, we optimize the loss function to pull positive features \( {f'}_{vq_k,_i}^+ \) closer to the positive anchor feature \( {f'}_{vq_a}^+ \) while pushing all \(M_2\) negative features \( {f'}_{vq_k,_j}^- \) away, as formulated in Eq. \ref{eq5}, where \( \tau^+ \) is the positive temperature.
\\On the other hand, for $\mathcal{L}_{NAC}$, we consider the negative features of the randomly selected tier as the negative anchor (${f'}_{vq_a}^-$). We calculate the cosine similarity between (${f'}_{vq_a}^-$, ${f'}_{vq_k,_i}^-$) and (${f'}_{vq_a}^-$, ${f'}_{vq_k,j}^+$), where $i$ and $j$ iterate over $M_2$ negative and $M_1$ positive features, respectively, while $k$ iterates over $K-1$ tiers, as stated in Eq. \ref{eq6}.
\begin{equation}
\small
\label{eq6}
\mathcal{{L}}_{NAC}=-\log \frac{\sum_{k=1}^{K-1} \sum_{i=1}^{M_2} \exp \left(sim({f'}_{vq_a}^-,{f'}_{vq_k,_i}^-)/ \tau^-\right)}{\sum_{k=1}^{K-1}  \sum_{j= 1}^{M_1} \exp \left(sim({f'}_{vq_a}^-,{f'}_{vq_k,_j}^+)/ \tau^-\right)}
\end{equation}
Finally, we optimize the loss to pull the negative features (${f'}_{vq_k,_i}^-$) closer to the negative anchor features (${f'}_{vq_a}^-$) while pushing all $M_1$ positive features (${f'}_{vq_k,_j}^+$) away, as shown in Eq. \ref{eq6}, where $\tau^-$ is the temperature parameter for $\mathcal{L}_{NAC}$.
The final loss $\mathcal{{L}}_{MTDA}$ is given in Eq. \ref{eq9} where ${w_p}$ and ${w_n}$ are tunable weights for $\mathcal{{L}}_{PAC}$ and $\mathcal{{L}}_{NAC}$ respectively.
\begin{equation}
\small
\label{eq9}
    \mathcal{L}_{MTDA} = {w_p} * \mathcal{{L}}_{PAC} + {w_n}* \mathcal{{L}}_{NAC}
\end{equation}

\noindent
\subsubsection{Temporal Uncertainty Aware Localization Loss}
The VQ-VCL task becomes more challenging when there is a high similarity between the frames belonging to different classes and the event boundaries are not well defined. This leads to noisy manual annotations. To reduce the effect of noisy annotations, we introduce a Temporal Uncertainty Aware Localization Loss ($\mathcal{L}_{URL}$). Instead of using binary ground truth, we generate two Gaussian distributions $T_s(x)$ and $T_e(x)$ corresponding to the true start frame ($t_s$) and true end frame ($t_e$) of the target video clip, with means $\mu_{s} = t_s$ and $\mu_{e} = t_e$  and standard deviation ($\sigma=1$) respectively as shown in Eq. \ref{eq10}.
\begin{equation}
\small
\label{eq10}
T_s(x) = \frac{1}{{\sigma \sqrt {2\pi } }}e^{{{ - \left( {x - \mu_s } \right)^2 } \mathord{\left/ {\vphantom {{ - \left( {x - \mu } \right)^2 } {2\sigma ^2 }}} \right. \kern-\nulldelimiterspace} {2\sigma ^2 }}},\ T_e(x) = \frac{1}{{\sigma \sqrt {2\pi } }}e^{{{ - \left( {x - \mu_e } \right)^2 } \mathord{\left/ {\vphantom {{ - \left( {x - \mu } \right)^2 } {2\sigma ^2 }}} \right. \kern-\nulldelimiterspace} {2\sigma ^2 }}}
\end{equation}
Finally, we optimize the KL-divergence loss between the predicted ($P_s(x)$, $P_e(x)$) and true ($T_s(x)$, $T_e(x)$)  start and end distribution and combine as shown in Eqs. \ref{eq12} and \ref{eq14} respectively.


\begin{equation}
\small
\label{eq12}
\begin{split}
& KL_{s}(P_s||T_s)=\sum_{x}P_s(x)\log(\frac{P_s(x)}{T_s(x)}) , \\ &\ KL_{e}(P_e||T_e)=\sum_{x}P_e(x)\log(\frac{P_e(x)}{T_e(x)})
\end{split}
\end{equation}



\begin{equation}
\small
\label{eq14}
\mathcal{L}_{URL} = KL_{s} + KL_{e}
\end{equation}
Finally, we combine  Eqs \ref{eq9} and \ref{eq14} to give the total loss $\mathcal{L}$ used to train our model as formulated in Eq. \ref{eq15}.
\begin{equation}
\small
\label{eq15}
\mathcal{L} = \mathcal{L}_{MTDA} +\mathcal{L}_{URL}
\end{equation}


\section{Experiments and Results}
\subsubsection{Dataset and Implementation}
We evaluated MCAT on two distinct fetal ultrasound video datasets following \cite{Mis_STANLOC_MICCAI2024} and one egocentric computer vision dataset, Ego4D VQ-VCL, which we created based on the Ego4D dataset \cite{grauman2022ego4d}.
The first dataset consists of fetal heart video sweeps from CAIFE (Development of
Clinical Artificial Intelligence Models in Fetal Echocardiogra-
phy for the Detection of Congenital Heart Defects). It includes 10-second transversal heart sweeps over the fetal heart (see Supp. Fig 3), scanning from the cardiac situs (Situs) to the four-chamber view (4CH), through the left ventricular outflow tract (LVOT), the three-vessel view (3VV), and finally, the three-vessel trachea view (3VT) of the fetal heart. Unlike routine heart scans, where the sonographer pauses to capture the perfect plane for each anatomical view, these sweeps continuously scan across the heart. The VQ-VCL task retrieves a standard heart-view clip given a visual query of the standard heart view. We used 200 healthy heart sweep videos for training and 47 videos for testing. The visual query for the heart sweep data consisted of 2804 standard frames extracted from 12 held-out videos. Further details about our unique heart sweep data are in Ultrasound dataset details section of Supplementary.
Our second dataset is derived from the PULSE \cite{drukker2021transforming} fetal ultrasound anomaly scan video dataset. We extracted clips for 8 fetal anatomical planes utilized for clinical anomaly detection, including Transventricular and Transcerebellar Views of the fetal head, Abdomen, Femur, and the 4CH, LVOT, 3VV, and 3VT views of the fetal heart. We trained the MCAT model on 200 videos and tested it on 30 videos. The visual query comprised 4378 standard frames extracted from 30 videos. 
As we introduce the VQ-VCL task, we acknowledge the lack of open-source datasets for model reproducibility. Therefore, we utilized the existing Ego4D \cite{grauman2022ego4d} dataset to create the Ego4D VQ-VCL dataset. We plan to release the dataset creation script along with our code to ensure the reproducibility of our work.
Video and visual query frames were resized to dimensions of \(224 \times 224\). During training, we augmented the dataset by sampling clips with varying start and end frames, each containing 150 frames. All models were trained for 200 epochs in PyTorch version 1.8 using a Tesla V100 32 GB GPU. We employed AdamW optimizer with a StepLR learning scheduler, utilizing cosine annealing with a step-size of 75. Our visual encoder was ResNet101, and both our multi-tier feature fusion transformers consisted of 2 layers each.


\begin{figure*}
    \centering
     \includegraphics[width=1.9\columnwidth]{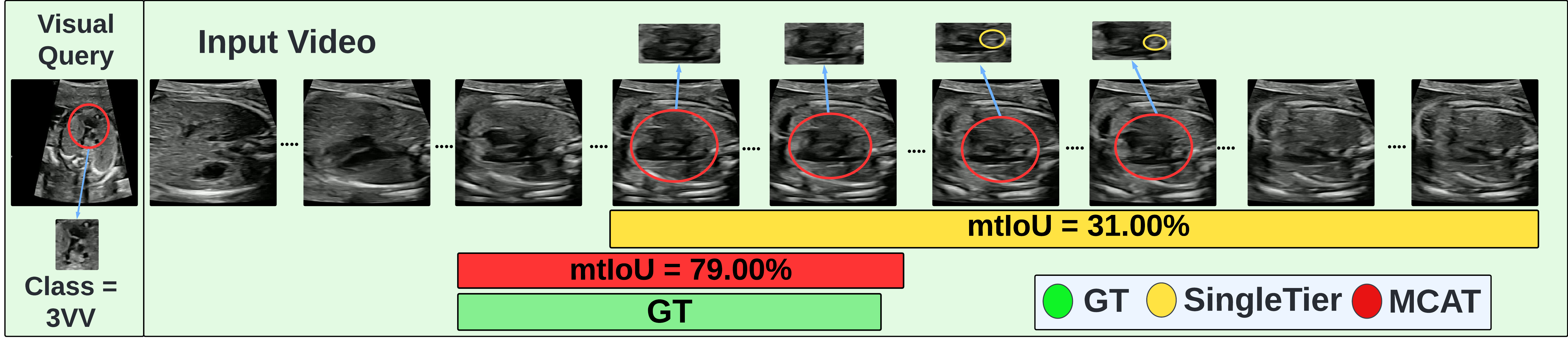}
    
\caption{This figure compares the predictions of a single-tier model with our multi-tier model for an LVOT visual query.
}
    \label{fig4}
\end{figure*}

\subsubsection{Metrics}To evaluate the performance of MCAT, we follow previous works on temporal video grounding \cite{wang2023protege,moon2023query} and our baselines \cite{goyal2023minotaur,jiang2023single}. Hence, we compute the mean temporal intersection-over-union (mtIoU) and "R @ t", where R represents recall measured at predefined temporal IoU (tIoU) thresholds (t). For our experiments, we report recall at thresholds \(t = 0.1, 0.3, 0.5\) and \(0.7\).

\begin{table}[htbp]
\centering
\scalebox{0.8}{ 
\begin{tabular}{c|c|c|c|c|c}
\toprule
\multicolumn{6}{c}{\textbf{Heart Sweep Data}} \\ \toprule
\textbf{Method} & \textbf{mtIoU} & \textbf{R@0.7} & \textbf{R@0.5} & \textbf{R@0.3} & \textbf{R@0.1} \\ \midrule
CS Sup CNN & $5.03$ & $0.00$ & $0.00$ & $4.00$ & $16.22$ \\ \midrule
\makecell{TubeDETR} & $12.72$ & $2.00$ & $2.00$ & $10.22$ & $20.00$ \\ \midrule
\makecell{MomentDETR} & $14.89$ & $0.00$ & $8.00$ & $25.00$ & $39.72$ \\ \midrule
\makecell{Resnet 3D} & $19.79$ & $6.00$ & $6.00$ & $23.22$ & $47.17$ \\ \midrule
\makecell{VQLOC} & $24.05$ & $2.50$ & $13.50$ & $34.50$ & ${62.17}$ \\ \midrule
\makecell{MCAT (Ours)} & $\mathbf{34.1}$ & $\mathbf{11.00}$ & $\mathbf{30.17}$ & $\mathbf{56.17}$ & $\mathbf{66.17}$ \\ \midrule

\multicolumn{6}{c}{\textbf{PULSE \cite{drukker2021transforming} Data}} \\ \midrule
\textbf{Method} & \textbf{mtIoU} & \textbf{R@0.7} & \textbf{R@0.5} & \textbf{R@0.3} & \textbf{R@0.1} \\ \midrule
CS Sup CNN & $2.6$ & $2.04$ & $2.04$ & $2.04$ & $2.04$ \\ \midrule
\makecell{TubeDETR} & $7.07$ & $4.76$ & $4.76$ & $4.76$ & $14.42$ \\ \midrule
\makecell{MomentDETR} & $10.34$ & $2.04$ & $6.93$ & $16.60$ & $21.50$ \\ \midrule
\makecell{Resnet 3D} & $17.89$ & $14.29$ & $17.14$ & $22.04$ & ${28.16}$ \\ \midrule
\makecell{VQLOC} & $12.62$ & $0.00$ & $14.29$ & $14.29$ & ${22.04}$ \\ \midrule
\makecell{MCAT (Ours)} & $\mathbf{30.63}$ & $\mathbf{26.80}$ & $\mathbf{31.70}$ & $\mathbf{34.56}$ & $\mathbf{39.32}$ \\\toprule

\multicolumn{6}{c}{\textbf{Ego4D \cite{grauman2022ego4d} VQ-VCL Dataset}} \\ \midrule
\textbf{Method} & \textbf{mtIoU} & \textbf{R@0.7} & \textbf{R@0.5} & \textbf{R@0.3} & \textbf{R@0.1} \\ \toprule
CS Sup CNN & $4.89$ & $0.00$ & $0.00$ & $7.93$ & $15.87$ \\\midrule
\makecell{Resnet 3D} & $10.72$ & $3.57$ & $8.33$ & $12.70$ & ${20.24}$ \\ \midrule
\makecell{VQLOC} & $25.35$ & $7.14$ & $19.44$ & $32.94$ & ${44.84}$ \\ \midrule
\makecell{MomentDETR} & $38.44$ & $15.08$ & $25.40$ & $52.78$ & $66.67$ \\ \midrule
\makecell{TubeDETR} & $38.59$ & $18.65$ & $32.94$ & $61.51$ & $71.03$ \\ \midrule
\makecell{MCAT (Ours)} & $\mathbf{43.94}$ & $\mathbf{32.54}$ & $\mathbf{39.68}$ & $\mathbf{64.68}$ & $\mathbf{71.83}$ \\ \bottomrule
\end{tabular}}
\caption{ \normalsize Quantitative comparison of MCAT }
\label{table1}
\end{table}

\subsubsection{Quantitative Results}
We compare, MCAT with ResNet3D CNN \cite{hara2017learning}, Cosine Similarity Supervised 2D CNN, TubeDETR \cite{yang2022tubedetr}, VQLOC \cite{jiang2023single}, and MomentDETR \cite{lei2021detecting}, as in Table \ref{table1}. Further details about baselines are in Supp.\\
From Table \ref{table1}, we observe that the cosine similarity supervised baseline performs worst, achieving an mtIoU of 5.03\%, R @ 0.7 = 0.00\% , R @ 0.5 = 0.0\% . This indicates the model's inability to effectively extract, fuse, and model long-range features from both the input video and the visual query. TubeDETR demonstrates improved performance with an mtIoU of 12.72\%, R @ 0.3 = 10.22\%. This improvement may be attributed to the spatio-temporal transformer in TubeDETR, facilitating the extraction and fusion of video features in both spatial and temporal dimensions. However, the model still struggles with longer interactions, as indicated by R @ 0.7 and R @ 0.3 both being 2.00\%, suggesting that the features from the visual query and the video are insufficient for modeling extended interactions. This limitation may be due to the direct concatenation of video and visual query features in the model. A similar pattern is observed with MomentDETR, where mtIoU is 14.89\%. The model handles short-range interactions well with R @ 0.3 = 25.00\% and R @ 0.1 = 39.72\%, but it performs poorly in capturing longer-range interactions (R @ 0.7 = 0.00\% and R @ 0.5 = 8.00\%), possibly due to the concatenation of visual query and video features. 
The ResNet3D baseline outperforms TubeDETR and MomentDETR, achieving an mtIoU of 19.79\% and demonstrating better modeling of longer interactions with R @ 0.7 = 6.00\% and R @ 0.5 = 6.00\%. ResNet3D also performs well in modeling shorter interactions with R @ 0.1 = 47.17\%. This improvement may be attributed to the equal interaction of the visual query with each frame of the video, achieved by concatenating them together for each frame. VQLOC achieves an mtIoU of 24.05\%, R @ 0.5 = 13.50\%, R @ 0.1 = 62.17\%, indicating its ability to model both short and long-range interactions.
MCAT outperforms all baselines with a mtIoU of 34.10\%,  10.05\% higher than VQLOC. Its performance in modeling long-range and short-range dependencies is significantly better, with R @ 0.7 = 11.00\%, R @ 0.5 = 30.17\%, R @ 0.3 = 56.17\%, and R @ 0.1 = 66.17\%. MCAT effectively models the relationship between the visual query and the video in both spatial and temporal dimensions due to the Multi-Tier Spatio-Temporal Fusion Transformer and disentanglement of class-specific spatio-temporal features through class-specific tokens. Additionally, the incorporation of boundary losses ($\mathcal{L}_{MTDA}$ and $\mathcal{L}_{URL}$) enhances its ability to detect boundaries, making it robust to noisy annotations and resulting in improved performance. A similar trend is seen in the PULSE \cite{drukker2021transforming} data (refer to Table \ref{table1}) and Ego4D VQ-VCL (refer to Table \ref{table1}), with our model MCAT outperforming the best-performing baseline by 12.74\% and 5.35\%, respectively. 
\subsubsection{Qualitative Results}
In the qualitative comparison, we analyze the single-tier model, which only utilizes features from the last layer (Tier=1), against our multi-tier model (Tiers=1, 2, 3). As shown in Fig. \ref{fig4}, the single-tier model struggles to differentiate between the 3VV and 3VT views in the video, as both views display three vessels. The critical distinction is the appearance of the trachea in the 3VT view, as highlighted by the yellow circle. Our multi-tier model successfully identifies this subtle change by leveraging features from multiple tiers, leading to significantly improved performance. Additional qualitative results are provided in the supplementary.

\subsection{Ablation Study}
This section reports ablation experiments to justify the inclusion of the key components in the MCAT model.

 \subsubsection{Importance of Tiers}
First we show the importance of Tiers to model performance. The model utilizing features only from the last layer of the visual backbone is referred to as Tier 1, while that utilising from Tier 1 and some layer before that is referred to as Tier 2, and so on. Experiments were performed for Tier = 1,2,3 where the feature sizes were $T \times 2048 \times 7 \times 7$, $T \times 1024 \times 14 \times 14$ and $T \times 512 \times 28 \times 28$ respectively. Table \ref{table_importance_of_tiers} shows that the model utilising only Tier 1 features performs the worst with mtIoU = 28.23 \%. This can be explained because Tier 1 features are insufficient to capture the fine-grained detail necessary to determine event boundaries and to distinguish between highly similar classes.
 When features from Tier 1 and Tier 2 are utilized, mtIoU increases by 3.1\%. The highest performance is seen with Tier 3, which surpasses the single Tier results by almost 6\% and Tier 2 by 2.77\% respectively stressing the advantage of using multi-Tier features to capture both low- and high-level details to distinguish fine-grained classes. 



\begin{table}[htbp]
\centering

\scalebox{0.8}{%
\begin{tabular}{c|c|c|c|c|c}
\toprule
\textbf{Tier} & \textbf{mtIoU} & \textbf{R@0.7} & \textbf{R@0.5} & \textbf{R@0.3} & \textbf{R@0.1} \\
\midrule
1       & 28.23  & 10.5   & 19.72  & 38.72  & \textbf{66.61} \\ \midrule
1, 2    & 31.33  & \textbf{13.0} & 28.22  & 42.44  & \textbf{66.61} \\ \midrule
1, 2, 3 & \textbf{34.10} & 11.00  & \textbf{30.17} & \textbf{56.17} & \textbf{66.17} \\
\bottomrule

\end{tabular}%
}
\caption{\normalsize Showing importance of multi-Tier features.}
\label{table_importance_of_tiers}
\end{table}

 \subsubsection{Importance of different Loss functions}
In Table \ref{table_importance_of_loss_functions}, we investigate the impact of each loss function on model performance. Our baseline loss is Cross-Entropy Loss, as shown in the first row of Table \ref{table_importance_of_loss_functions}. Replacing it with $\mathcal{L}_{URL}$ improved performance by 14\% mtIoU, highlighting the significance of incorporating uncertainty in loss functions when the ground truth annotation contains a high degree of noise. Additionally, we ablate the Multi-Tier, Dual Anchor Contrastive loss ($\mathcal{L}_{MTDA}$). Including this loss, which consists of our dual-anchor losses ($\mathcal{L}_{PAC}$ and $\mathcal{L}_{NAC}$), alongside $\mathcal{L}_{URL}$, further enhances performance. Specifically, mtIoU increases by 4.46\%, R @ 0.5 by 3.95\%, R @ 0.3 by 11.23\%, and R @ 0.1 by 9.78\%. These improvements indicate the importance of both positive and negative anchors to distinguish highly similar classes and to reduce confusion at event boundaries.


\begin{table}[htbp]
\centering

\scalebox{0.8}{%
\begin{tabular}{c|c|c|c|c|c|c}
\toprule
\textbf{$\mathcal{L}_{URL}$} & \textbf{$\mathcal{L}_{MTDA}$} & \textbf{mtIoU} & \textbf{R@0.7} & \textbf{R@0.5} & \textbf{R@0.3} & \textbf{R@0.1} \\
\midrule
\xmark & \xmark & 15.93 & 2.00 & 12.00 & 22.22 & 42.44 \\
\midrule
\checkmark & \xmark & 29.64 & 11.50 & 26.22 & 44.94 & 56.39 \\
\midrule
\checkmark & \checkmark & \textbf{34.10} & \textbf{11.00} & \textbf{30.17} & \textbf{56.17} & \textbf{66.17} \\
\bottomrule
\end{tabular}%
}
\caption{Analysis of contribution of different loss functions.}
\label{table_importance_of_loss_functions}
\end{table}

 \subsubsection{Sequential vs Parallel Fusion} In existing works utilizing multi-scale features \cite{wang2021pyramid,fan2021multiscale}, sequential feature fusion is employed. In sequential fusion, multi-scale features are projected into a common feature space and fused sequentially from coarse to fine.
In contrast, our work employs parallel fusion in the encoder to exploit the implicit bias of image/video modalities and capture minute variations. In parallel fusion, features from the video and visual query at each tier are fused separately, maintaining the original resolution across tiers. As shown in Table \ref{table_sequential_vs_parallel_fusion}, parallel fusion outperforms sequential fusion, improving R@0.3 by 11.45\%, R@0.5 by 6.67\%, and mtIoU by 2.33\%, demonstrating its superiority in capturing short-term and long-term interactions between the video and visual query.
\begin{table}[hbtp]
\centering
\scalebox{0.75}{%
\begin{tabular}{c|c|c|c|c|c}
\toprule
\textbf{Method} & \textbf{mtIoU} & \textbf{R@0.7} & \textbf{R@0.5} & \textbf{R@0.3} & \textbf{R@0.1} \\
\midrule
Sequential Fusion & 31.77 & \textbf{11.00} & 23.50 & 44.72 & \textbf{68.39} \\
\midrule
Parallel Fusion (Ours) & \textbf{34.10} & \textbf{11.00} & \textbf{30.17} & \textbf{56.17} & 66.17 \\
\bottomrule
\end{tabular}%
}
\caption{ Effect of sequential and parallel fusion.}
\label{table_sequential_vs_parallel_fusion}
\end{table}

\subsubsection{ Video Query Fusion }We examined the impact of  Concat Self-Attention (SA) method, which is popular for multi-modality fusion~\cite{yang2022tubedetr,devlin2018bert,lei2021detecting}, with Cross-Attention feature fusion for fusing visual query features with video features. As shown in Table \ref{table_video_query_fusion}, cross-attention fusion significantly outperforms Concat SA fusion by 20.63\% mtIoU. This improvement is because, in cross-attention fusion, the Key/Value is derived from the visual query while the Query comes from the video. This ensures a substantial contribution from the visual query features, resulting in query-aware fused representations. In contrast, Concat SA involves directly concatenating the visual query features with the video features and performing self-attention on the entire sequence. This approach reduces the contribution of visual query features in the fused representation, as the VQ feature becomes just one element within the sequence.
\begin{table}[hbtp]
\centering

\scalebox{0.75}{%
\begin{tabular}{c|c|c|c|c|c}
\toprule
\textbf{Method} & \textbf{mtIoU} & \textbf{R@0.7} & \textbf{R@0.5} & \textbf{R@0.3} & \textbf{R@0.1} \\
\midrule
Concat Self-Attention & 14.47 & 6.00 & 8.00 & 20.00 & \textbf{34.22} \\
\midrule
Parallel Fusion (Ours) & \textbf{34.10} & \textbf{11.00} & \textbf{30.17} & \textbf{56.17} & \textbf{66.17} \\
\bottomrule
\end{tabular}%
}
\caption{Comparing methods for video-visual query fusion.}
\label{table_video_query_fusion}
\end{table}

 \noindent \subsubsection{ Class-Specific Tokens vs Generic Embedding} In existing works such as \cite{yang2022tubedetr,jiang2023single}, the temporal transformer learns a generic embedding that is shared across classes and corresponds to the number of frames in the video. While this approach may be effective for coarse-grained videos, it results in sub-optimal performance for fine-grained videos. As shown in Table \ref{table_class_specific_tokens_vs_generic_embedding}, our class-specific embedding outperforms the generic embedding by 3.1\% with 96\% fewer tokens.
\begin{table}[hbtp]
\centering

\scalebox{0.68}{%
\begin{tabular}{c|c|c|c|c|c}
\toprule
\textbf{Method} & \textbf{mtIoU} & \textbf{R@0.7} & \textbf{R@0.5} & \textbf{R@0.3} & \textbf{R@0.1} \\
\midrule
Generic Embedding & 31.00 & \textbf{13.22} & \textbf{27.72} & 40.44 & 65.67 \\
\midrule
Class-Specific Embedding (Ours) & \textbf{34.10} & 11.00 & \textbf{30.17} & \textbf{56.17} & \textbf{66.17} \\
\bottomrule
\end{tabular}%
}
\caption{Generic Embedding vs Class-Specific Token}
\label{table_class_specific_tokens_vs_generic_embedding}
\end{table}

  \subsubsection{Tier-Specific Embedding}
We demonstrate the importance of using scale-specific embedding in the Query-Guided Spatial Transformer. As shown in Table \ref{table_tier_specific_embedding}, Tier-Specific Embedding improves model performance by 3.52\%, highlighting its crucial role in capturing tier-specific features essential for fine-grained video retrieval.
\begin{table}[hbtp]
\centering

\scalebox{0.7}{%
\begin{tabular}{c|c|c|c|c|c}
\toprule
\textbf{Method} & \textbf{mtIoU} & \textbf{R@0.7} & \textbf{R@0.5} & \textbf{R@0.3} & \textbf{R@0.1} \\
\midrule
W/O Tier-Specific Embedding & 30.58 & \textbf{15.50} & 25.72 & 42.67 & 62.61 \\
\midrule
W/ Tier-Specific Embedding & \textbf{34.10} & 11.00 & \textbf{30.17} & \textbf{56.17} & \textbf{66.17} \\
\bottomrule
\end{tabular}%
}
\caption{Importance of Tier-Specific Embedding}
\label{table_tier_specific_embedding}
\end{table}



\section{Conclusion}
This paper introduces an visual-query based solution for detecting standard anatomy video clips in fetal US videos.
Our model MCAT, is a video-based transformer that leverages multi-tier features and class-specific token learning to understand the video with the visual query. This significantly improves video-clip localization compared to models that use single-tier features with 96\% less tokens.This enables the model to retrieve the relevant video clip based on a visual query in just 2.69 seconds while using only 4.62 GB of memory during inference, allowing it to run effectively on affordable GPUs, even in resource-limited settings. Additionally, we introduce a temporal uncertainty-aware loss to handle the inherent noise in real-world annotations. 
Furthermore, to differentiate highly similar classes in fine-grained videos, we propose a contrastive loss that utilizes multi-tier features and multi-anchor guidance to learn subtle class-discriminative features. 
We apply MCAT to real-world standard plane video-clip detection task with limited data and fine-grained classes and validate its effectiveness through comparisons with SOTA baselines, demonstrating significant improvements in localization accuracy and resource efficiency. These traits make our model beneficial for prenatal care in LMICs, where access to advanced diagnostic tools and skilled health professionals is limited.
\section{Acknowledgments}
This work was supported in part by the InnoHK-funded Hong Kong Centre for Cerebro-cardiovascular Health Engineering (COCHE) Project 2.1 (Cardiovascular risks in early life and fetal echocardiography), the UK EPSRC (Engineering and Physical Research Council) Programme Grant EP/T028572/1 (VisualAI), and a UK EPSRC Doctoral
Training Partnership award, and the UKRI grant EP/X040186/1
(Turing AI Fellowship).

\bibliography{aaai25}

\end{document}